\documentclass[twoside]{article}
\usepackage[accepted]{aistats2018}

\usepackage[authoryear,round]{natbib}
\usepackage{graphicx}
\usepackage{color}
\usepackage{url}
\usepackage{listings}
\usepackage{caption}
\usepackage{subcaption}

\usepackage{amsmath}
\usepackage{amsthm}
\usepackage{amssymb}
\usepackage{amsfonts}

\usepackage{times}

\usepackage{hyperref}
\newcommand{\pushleft}[1]{\ifmeasuring@#1\else\omit$\displaystyle#1$\hfill\fi\ignorespaces}

\newcommand{\R}{{\mathbb{R}}}
\newcommand{\p}{{\mathsf{P}}}

\newcommand{\iid}{\overset{\mathrm{iid}}{\sim}}

\newcommand{\embeddingp}{{\mu_{\p}}}

\newcommand{\domX}{{\mathcal{X}}}

\newcommand{\rkhs}{{\mathcal{H}_k}}

\DeclareMathOperator{\diag}{\mathrm{diag}}
\DeclareMathOperator{\GP}{\mathcal{GP}}
\DeclareMathOperator{\E}{\mathbb E}
\DeclareMathOperator{\N}{\mathcal N}
\DeclareMathOperator*{\argmin}{argmin}
\newcommand{\tp}{^\mathsf{T}}
\DeclareMathOperator{\Var}{Var}

\newcommand{\httpurl}[1]{\href{http://#1}{\nolinkurl{#1}}}

\begin{document}
\lstset{
  basicstyle=\ttfamily,
  columns=fullflexible,
  keepspaces=true,
}

\twocolumn[

\aistatstitle{Bayesian Approaches to Distribution Regression}

\aistatsauthor{Ho Chung Leon Law$^\ast$ \And Danica J.\ Sutherland$^\ast$ \And  Dino Sejdinovic \And Seth Flaxman}
\aistatsaddress{
     University of Oxford \\
     ho.law@spc.ox.ac.uk
\And %
     University College London \\
     djs@djsutherland.ml
\And %
     University of Oxford \\
     dino.sejdinovic@stats.ox.ac.uk
\And %
     Imperial College London \\
     s.flaxman@imperial.ac.uk
}
]

\begin{abstract}
Distribution regression has recently attracted much interest as a generic solution to the problem of supervised learning where labels are available at the group level, rather than at the individual level. Current approaches, however, do not propagate the uncertainty in observations due to sampling variability in the groups. This effectively assumes that small and large groups are estimated equally well, and should have equal weight in the final regression. We account for this uncertainty with a Bayesian distribution regression formalism, improving the robustness and performance of the model when group sizes vary. We frame our models in a neural network style, allowing for simple MAP inference using backpropagation to learn the parameters, as well as MCMC-based inference which can fully propagate uncertainty. We demonstrate our approach on illustrative toy datasets, as well as on a challenging problem of predicting age from images.
\end{abstract}

\section{INTRODUCTION}
Distribution regression is the problem of learning a regression function from samples of a distribution to a single set-level label. For example, we might attempt to infer the sentiment of texts based on word-level features, to predict the label of an image based on small patches,
or even perform traditional parametric statistical inference by learning a function from sets of samples to the parameter values.

Recent years have seen wide-ranging applications of this framework,
including
inferring summary statistics in Approximate Bayesian Computation \citep{MitSejTeh2016},
estimating Expectation Propagation messages \citep{JitGreHeeEslLakSejSza2015},
predicting the voting behaviour of demographic groups \citep{flaxman2015ecological,flaxman2016understanding},
and learning the total mass of dark matter halos from observable galaxy velocities \citep{Ntampaka2015,Ntampaka2016}.
Closely related distribution classification problems also include identifying the direction of causal relationships from data \citep{Lopez-paz2015}
and classifying text based on bags of word vectors \citep{Yoshikawa2014,Kusner2015}.

One particularly appealing approach to the distribution regression problem
is to represent the input set of samples by their kernel mean embedding (described in Section \ref{sec:overview}),
where distributions are represented as single points in a reproducing kernel Hilbert space.
Standard kernel methods can then be applied for distribution regression, classification, anomaly detection, and so on.
This approach was perhaps first popularized by \citet{muandet:smm};
\citet{szabo2015two} provided a recent learning-theoretic analysis.

\renewcommand{\thefootnote}{\fnsymbol{footnote}}%
\footnotetext[1]{These authors contributed equally.}%
\renewcommand{\thefootnote}{\arabic{footnote}}%
In this framework, however, each distribution is simply represented by the empirical mean embedding,
ignoring the fact that large sample sets are much more precisely understood than small ones.
Most studies also use point estimates for their regressions, such as kernel ridge regression or support vector machines, thus ignoring uncertainty both in the distribution embeddings and in the regression model.

\paragraph{Our Contributions}
We propose a set of Bayesian approaches to distribution regression.
The simplest method, similar to that of \cite{flaxman2015ecological}, is to use point estimates of the input embeddings but account for uncertainty in the regression model with simple Bayesian linear regression.
Alternatively, we can treat uncertainty in the input embeddings but ignore model uncertainty with the proposed Bayesian mean shrinkage model, which builds on a recently proposed Bayesian nonparametric model of uncertainty in kernel mean embeddings \citep{flaxman2016bayesian}, and then use a sparse representation of the desired function in the RKHS for prediction in the regression model.
This model allows for a full account of uncertainty in the mean embedding, but requires a point estimate of the regression function for conjugacy; we thus use backpropagation to obtain a MAP estimate for it as well as various hyperparameters.
We then combine the treatment of the two sources of uncertainty into a fully Bayesian model and use Hamiltonian Monte Carlo for efficient inference.
Depending on the inferential goals, each model can be useful.
We demonstrate our approaches on an illustrative toy problem as well as a challenging real-world age estimation task.

\section{BACKGROUND}
\label{section:background}
\subsection{Problem Overview} \label{sec:overview}
Distribution regression is the task of learning a classifier or a regression
function that maps probability distributions to labels. The challenge of
distribution regression goes beyond the standard supervised learning setting: we do not
have access to exact input-output pairs since the true inputs,
probability distributions,
are observed only through samples from that distribution:
\begin{equation}
\left(\{x_j^1\}_{j=1}^{N_1}, y_1\right),
\ldots,
\left(\{x_j^n\}_{j=1}^{N_n}, y_n\right)
\label{eq:dist-dataset}
,\end{equation}
so that each bag $\{x_j^i\}_{j=1}^{N_i}$ has a label $y_i$ along with $N_i$ individual
observations $x_j^i \in \domX$.
We assume that the observations $\{ x_j^i \}_{j=1}^{N_i}$ are i.i.d.\ samples from some unobserved distribution $\p_i$,
and that the true label $y_i$ depends only on $\p_i$.
We wish to avoid making any strong parametric assumptions on the $\p_i$.
For the present work, we will assume the labels $y_i$ are real-valued;
Appendix \ref{app:binary} shows an extension to binary classification.
We typically take the observation space $\domX$ to be a subset of $\R^p$,
but it could easily be a structured domain such as text or images,
since we access it only through a kernel \citep[for examples, see e.g.][]{gartnerstructured}.

We consider the standard approach to distribution regression, which relies on
kernel mean embeddings and kernel ridge regression.  For any positive definite
kernel function $k:\domX\times\domX\to\R$, there exists a unique reproducing
kernel Hilbert space (RKHS) $\rkhs$, a possibly infinite-dimensional space of
functions $f:\domX\to\R$ where evaluation can be written as an inner product,
and in particular $f(x)=\langle f,k(\cdot,x) \rangle_\rkhs$ for all $f\in\rkhs,
x\in\domX$.
Here $k(\cdot, x) \in \rkhs$ is a function of one argument, $y \mapsto k(y, x)$.

Given a probability measure $\p$ on $\domX$, let us define the kernel mean embedding into $\rkhs$ as
\begin{equation}
   \embeddingp =\int k\left(\cdot,x\right) \p(dx) \in \rkhs.
 \label{eq:kme2}
\end{equation}
Notice that $\embeddingp$ serves as a high- or infinite-dimensional vector representation of $\p$.
For the kernel mean embedding of $\p$ into $\rkhs$ to be well-defined, it suffices that $\int \sqrt{k(x,x)}\p(dx)<\infty$,
which is trivially satisfied for all $\p$ if $k$ is bounded.
Analogously to the reproducing property of RKHS, $\embeddingp$ represents the expectation function on $\rkhs$:
$\int h(x)\p(dx)=\langle h,\embeddingp \rangle_\rkhs$.
For so-called \emph{characteristic} kernels \citep{sriperumbudur2010hilbert}, every probability measure has a unique embedding,
and thus $\embeddingp$ completely determines the corresponding probability measure.

\subsection{Estimating Mean Embeddings}
\label{section:kme:estimators}
For a set of samples $\{x_j\}^n_{j=1}$ drawn iid from $\p$, the empirical estimator of $\embeddingp$, $\widehat{\embeddingp} \in \rkhs$, is given by
\begin{equation}
	\widehat{\embeddingp} = \mu_{\widehat\p} =
	\int k\left(\cdot,x\right) \hat\p(dx) =
	\frac{1}{n}\sum_{j=1}^n k(\cdot,x_j).
\label{eq:kme-estimator}
\end{equation}
This is the standard estimator used by previous distribution regression approaches,
which the reproducing property of $\rkhs$ shows us
corresponds to the kernel
\begin{equation}
	\langle \widehat{\embeddingp_i}, \widehat{\embeddingp_j} \rangle_\rkhs
	= \frac{1}{N_i N_j} \sum_{\ell=1}^{N_i} \sum_{r=1}^{N_j} k(x_\ell^i, x_r^j)
\label{eq:kme-kernel}
.\end{equation}
But \eqref{eq:kme-estimator} is an empirical mean estimator in a high- or infinite-dimensional space,
and is thus subject to the well-known \emph{Stein phenomenon},
so that its performance is dominated by the James-Stein shrinkage estimators.
Indeed, \citet{muandet2014kernel} studied shrinkage estimators for mean embeddings,
which can result in substantially improved performance for some tasks \citep{Ramdas2015}.
\citet{flaxman2016bayesian} proposed a Bayesian analogue of shrinkage estimators, which we now review.

This approach consists of (1) a Gaussian Process prior $\mu_\p\sim \GP(m_0, r(\cdot, \cdot))$ on $\rkhs$, where $r$ is selected to ensure that $\embeddingp \in \rkhs$ almost surely and (2) a normal likelihood $\widehat{\embeddingp}({\bf x}) \mid \embeddingp({\bf x})\sim \mathcal N(\embeddingp({\bf x}),\Sigma)$. Here, conjugacy of the prior and the likelihood leads to a Gaussian process posterior on the true embedding $\embeddingp$,
given that we have observed $\widehat{\embeddingp}$ at some set of locations ${\bf x}$.
The posterior mean is then essentially identical to a particular shrinkage estimator of \citet{muandet2014kernel}, but the method described here has the extra advantage of a closed form uncertainty estimate, which we utilise in our distributional approach. For the choice of $r$, we use a Gaussian RBF kernel $k$, and choose either $r = k$ or, following \citet{flaxman2016bayesian}, $r(x, x') = \int k(x,z) \, k(z,x') \, \nu(dz)$ where $\nu$ is proportional to a Gaussian measure. For details of our choices, and why they are sufficient for our purposes, see Appendix \ref{r_choice}.

This model accounts for the uncertainty based on the number of samples $N_i$, shrinking the embeddings for small sample sizes more. As we will see, this is essential in the context of distribution regression, particularly when bag sizes are imbalanced.

\subsection{Standard Approaches to Distribution Regression} \label{sec:standard-dist-reg}
\begin{figure}
    \centering
    \includegraphics[width=.35\paperwidth]{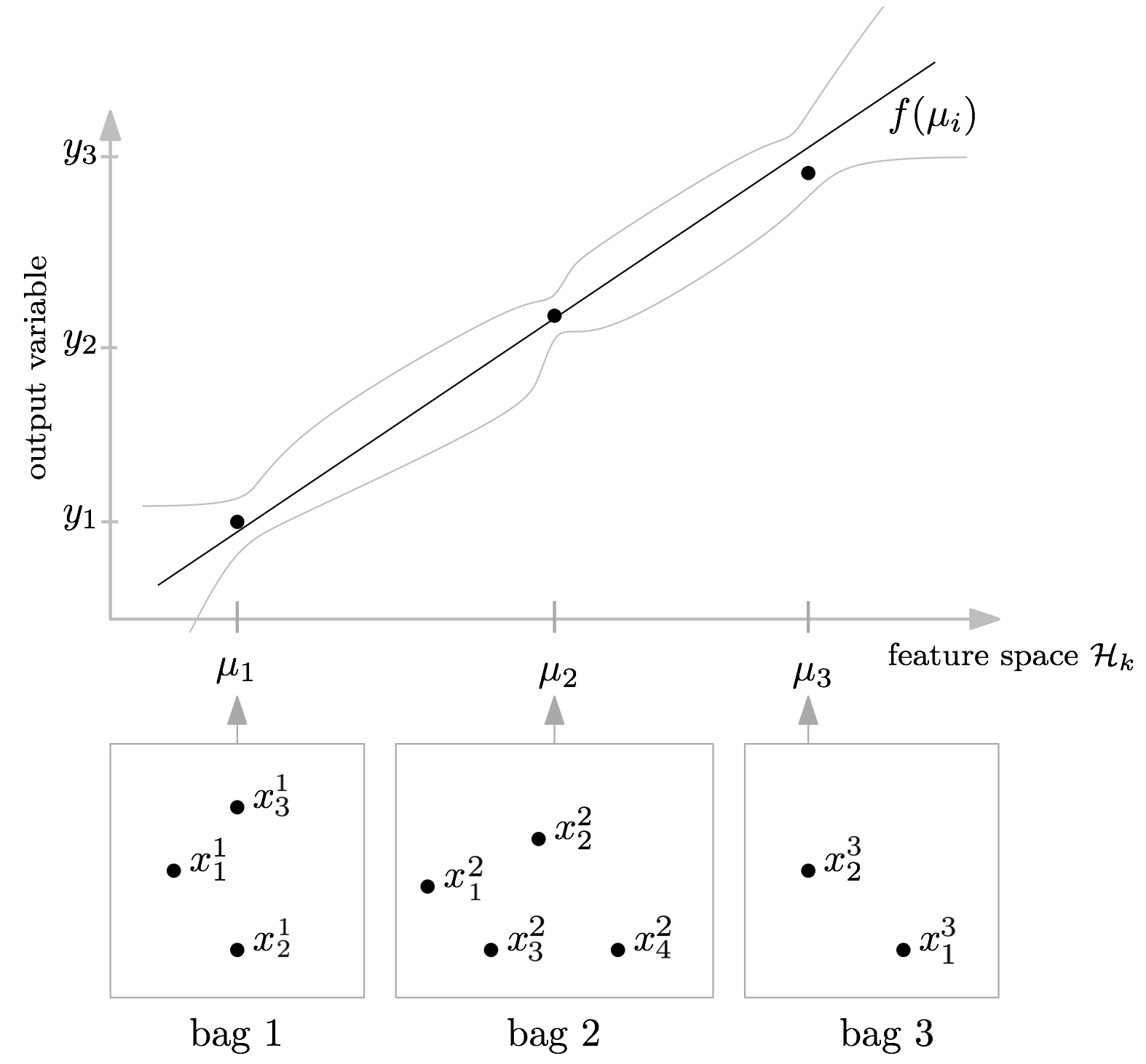} %
	\caption{Each bag is summarised by a kernel mean embedding $\mu_i \in \mathcal{H}_k$; a regression function $f: \mathcal{H}_k \rightarrow \mathbb{R}$ predicts labels $y_i \in \mathbb{R}$. We propose a Bayesian approach to propagate uncertainty due to the number of samples in each bag, obtaining posterior credible intervals illustrated in grey.}
    \label{fig:illustration}
\end{figure}

Following \cite{szabo2015two}, assume that the probability distributions $\p_i$ are
each drawn randomly from some unknown meta-distribution over probability
distributions, and take a two-stage approach, illustrated as in Figure~\ref{fig:illustration}.
Denoting the feature map $k(\cdot, x) \in \rkhs$ by $\phi(x)$,
one uses the empirical kernel mean estimator \eqref{eq:kme-estimator}
to separately estimate the mean of each group:
\begin{equation}
\widehat{\mu_1} = \frac{1}{N_1}\sum_{j=1}^{N_1} \phi(x_j^1), ~~ \ldots, ~~ \widehat{\mu_n} = \frac{1}{N_n} \sum_{i=1}^{N_n} \phi(x^n_j)
\label{eq:mean-estimator}
.\end{equation}
Next, one uses kernel ridge regression \citep{saunders1998ridge} to
learn a function $f : \rkhs \to \R$,
by minimizing the squared loss with an RKHS complexity penalty:
\begin{equation*}
\hat f = \argmin_{f \in \mathcal{H}_K} \sum_i (y_i - f(\widehat{\mu_i}))^2 + \lambda \|f\|^2_{\mathcal{H}_K}
.\end{equation*}
Here $K : \rkhs \times \rkhs \to \R$ is a ``second-level'' kernel on mean embeddings.
If $K$ is a linear kernel on the RKHS $\rkhs$, then the resulting method can be interpreted as a linear (ridge) regression on mean embeddings, which are themselves nonlinear transformations of the inputs.
A nonlinear second-level kernel on $\rkhs$ sometimes improves performance \citep{muandet:smm,szabo2015two}.

Distribution regression as described is not scalable for even modestly-sized datasets,
as computing each of the $\mathcal{O}(n^2)$ entries of the relevant kernel matrix requires time $\mathcal{O}(N_i N_j)$.
Many applications have thus used variants of random Fourier features \citep{rahimi2007random}.
In this paper we instead expand in terms of landmark points drawn randomly from the observations,
yielding radial basis networks \citep{broomhead1988radial}
with mean pooling. %

{

\section{MODELS}
\label{section:model}
We consider here three different Bayesian models, with each model encoding different types of uncertainty.
We begin with a non-Bayesian RBF
network formulation of the standard approach to distribution regression as a baseline, before
refining this approach to better propagate uncertainty in bag size, as well as model parameters.

}\begin{figure}
    \centering

    \includegraphics[width=.35\paperwidth]{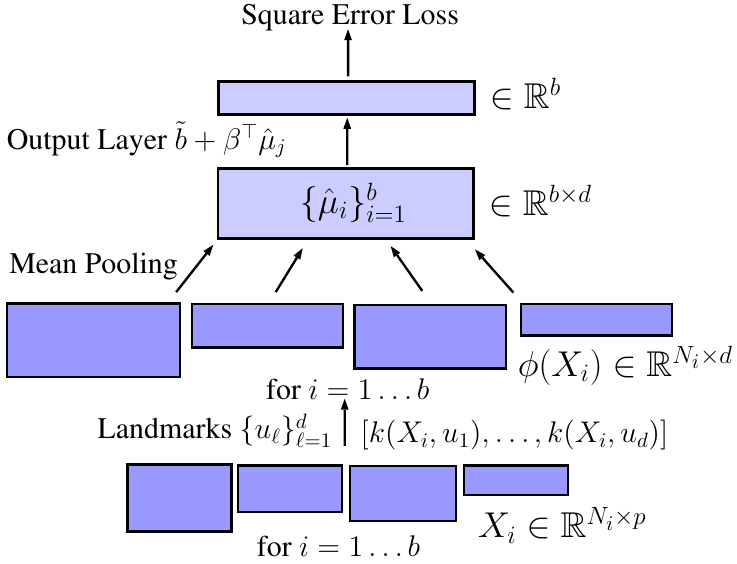}
        \caption{Our baseline model, a RBF network for distribution regression. $X_i$ represents the matrix of samples for bag $i$, while $k(X_i, u_\ell)$ represents the element wise operation on each row of $X_i$, with $b$ representing the batch size for stochastic gradient descent.}
    \label{fig:model-rbfnet}
\end{figure}

\subsection{Baseline Model} \label{section:baseline-model}
The baseline RBF network formulation we employ here is a variation of the approaches of \citet{broomhead1988radial}, \citet{Que2016}, \citet{law2017testing}, and \citet{zaheer2017deep}.
As shown in Figure \ref{fig:model-rbfnet},
the initial input is a minibatch consisting of several bags $X_i$, each containing $N_i$ points.
Each point is then converted to an explicit featurisation,
taking the role of $\phi$ in \eqref{eq:mean-estimator},
by a radial basis layer:
$x^i_j \in \R^p$ is mapped to
\[ \phi(x^i_{j}) = [k(x^i_{j},u_1), \ldots, k(x^i_{j},u_d)]^{\top} \in \mathbb{R}^d \]
where $\mathbf{u}= \{ u_\ell \}_{\ell=1}^d$ are landmark points.
A mean pooling layer yields the estimated mean embedding $\hat{\mu}_i$ corresponding to each of
the bags $j$ represented in the minibatch,
where $\hat{\mu}_i = \frac{1}{N_i} \sum_{j=1}^{N_i} \phi(x^i_j)$.\footnote{In the implementation, we stack all of the bags $X_i$ into a single matrix of size $\sum_j N_j \times d$ for the first layer, then perform pooling via sparse matrix multiplication.}
Finally, a fully connected output layer gives
real-valued labels $\hat y_i = \beta\tp \hat\mu_i + b$. As a loss function we use the mean square error $\frac1n \sum_i (\hat y_i - y_i)^2$.
For learning, we use backpropagation with the Adam optimizer \citep{Kingma2015}.
To regularise the network, we use early stopping on a validation set, as well as an $L_2$ penalty corresponding to a normal prior on $\beta$.

\begin{figure}
    \centering
    \includegraphics[width=.33\paperwidth]{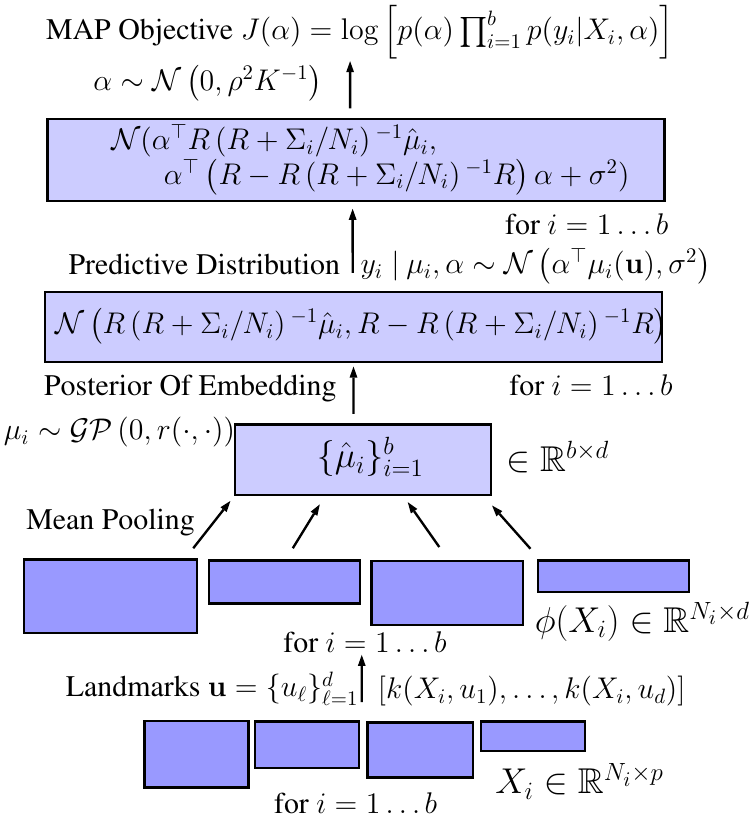}
        \caption{Our Bayesian mean shrinkage pooling model. This diagram takes $m_0 = \mathbf{0}$, $\eta =1$ and $\mathbf{u} = \mathbf{z}$, so that $R = R_{\mathbf{z}} = R_{\mathbf{zz}}$, and $K_\mathbf{z} = K$.}
    \label{fig:model-shrinkage}
\end{figure}

\subsection{Bayesian Linear Regression Model}
\label{section:blr}
The most obvious approach to adding uncertainty to the model of Section \ref{section:baseline-model} is to encode uncertainty over regression parameters $\beta$ only, as follows:
\begin{equation*}
\beta \sim \mathcal{N}(0, \rho^2)
    \qquad
y_{i} \mid \mathbf x_i, \beta \sim \N(\beta\tp \hat\mu_i, \sigma^2)
.\end{equation*}
This is essentially Bayesian linear regression on the empirical mean embeddings,
and is closely related to the model of \citet{flaxman2015ecological}.
Here, we are working directly with the finite-dimensional $\hat{\mu}_i$, unlike the infinite-dimensional $\mu_i$ before. Due to the conjugacy of the model, we can easily obtain the predictive distribution $y_i \mid \bf{x_i}$, integrating out the uncertainty over $\beta$.
This provides us with uncertainty intervals for the predictions $y_i$.

For model tuning, we can maximise the model evidence, i.e.\ the marginal log-likelihood (see \cite{bishop2006} for details), and use backpropagation through the network to learn $\sigma$ and $\rho$ and any kernel parameters of interest.\footnote{Note that unlike the other models considered in this paper, we cannot easily do minibatch stochastic gradient descent, as the marginal log-likelihood does not decompose for each individual data point.}

\subsection{Bayesian Mean Shrinkage Model}
\label{section:mean-shrinkage}
A shortcoming of the prior models, and of the standard approach in
\cite{szabo2015two}, is that they ignore uncertainty in the first level of
estimation due to varying number of samples in each bag. Ideally we would
estimate not just the mean embedding per bag, but also a measure of the
sample variance, in order to propagate this information regarding uncertainty from the bag size through the model.
Bayesian tools provide a natural framework for this problem.

We can use the Bayesian nonparametric prior over kernel
mean embeddings \citep{flaxman2016bayesian} described in Section
\ref{section:kme:estimators}, and observe the empirical embeddings at the landmark points $\mathbf{u_i}$.
For $\mathbf{u_i}$, we take a fixed set of landmarks, which we can choose via $k$-means clustering or sample without replacement \citep{Que2016}. Using the conjugacy of the model to the Gaussian process prior $\mu_i \sim \GP ( m_0, \eta r(.,.))$, we obtain a closed-form posterior Gaussian process whose evaluation at points $\mathbf{h} = \{h_s\}_{s=1}^{n_h}$ is:
\begin{multline*} %
\mu_{i}(\mathbf{h}) \mid \mathbf{x_i} \sim
   \mathcal{N}\left(
     R_{\mathbf{h}}\left(R+\Sigma_i/N_i\right)^{-1} (\hat{\mu}_{i} - m_0) + m_0,
 \right.\\\left.
 	 R_{\mathbf{h}\mathbf{h}} - R_\mathbf{h}\left( R + \Sigma_i / N_{i} \right)^{-1} R_\mathbf{h}^\top \right)
\end{multline*}
where $R_{st} = \eta r(u_s, u_t), (R_{\mathbf{h}\mathbf{h}})_{st}=\eta r(h_s, h_t), (R_\mathbf{h})_{st} =  \eta r(h_s, u_t)$,
and $\mathbf{x_i}$ denotes the set $\{x_j^i\}_{j=1}^{N_i}$.
We take the prior mean $m_0$ to be the average of the $\hat{\mu}_i$; under a linear kernel $K$, this means we shrink predictions towards the mean prediction.
Note $\eta$ essentially controls the strength of the shrinkage: a smaller $\eta$ means we shrink more strongly towards $m_0$. We take $\Sigma_i$ to be the average of the empirical covariance of $\{ \varphi(x^i_j) \}_{j=1}^{N_i}$ across all bags, to avoid poor estimation of $\Sigma_i$ for smaller bags. More intuition about the behaviour of this estimator can be found in Appendix \ref{app:shrink}.

Now, supposing we have normal observation error $\sigma^2$, and use a linear kernel as our second level kernel $K$, we have:
\begin{equation}
y_{i}\;|\:\mu_{i}, f  \sim  \mathcal{N}\left( \langle f , \mu_{i} \rangle_{\mathcal{H}_k} ,\sigma^{2}\right)
\end{equation}
where $f \in \mathcal{H}_k$. Clearly, this is difficult to work with; hence we parameterise $f$ as $f = \sum_{\ell=1}^s \alpha_\ell k(\cdot, z_\ell)$, where $\mathbf{z} = \{ z_\ell \}^s_{\ell=1}$ is a set of landmark points for $f$, which we can learn or fix. (Appendix \ref{app:representer} gives a motivation for this approximation using the representer theorem.)  Using the reproducing property, our likelihood model becomes:
\begin{equation}
y_{i}\;|\:\mu_{i}, \alpha  \sim  \N\left(\alpha\tp \mu_i(\mathbf{z}) ,\sigma^{2}\right)
\end{equation}
where $ \mu_i(\mathbf{z}) = [\mu_i(z_1), \dots, \mu_i(z_s)]^\top$.
For fixed $\alpha$ and $\mathbf{z}$ we can analytically integrate out the dependence on $\mu_i$,
and the predictive distribution of a bag label becomes
\begin{align*}
    y_i \mid \mathbf x_i, \alpha &\sim \N(\xi_i^\alpha, \nu_i^\alpha) \\
    \xi_i^\alpha
    &= \alpha^{\top}R_\mathbf{z}\left(R + \frac{\Sigma_i}{N_i}\right){}^{-1}(\hat{\mu}_i - m_0) + \alpha\tp m_0
\\
    \nu_i^\alpha
    &= \alpha\tp \left( R_\mathbf{zz} - R_\mathbf{z} \left(R + \frac{\Sigma_i}{N_i} \right)^{-1} R_\mathbf{z}\tp \right) \alpha + \sigma^2
.\end{align*}
The prior $\alpha \sim \mathcal{N}(0, \rho^2 K_{\mathbf{z}}^{-1})$,
where $K_{\mathbf{z}}$ is the kernel matrix on $\mathbf{z}$,
gives the standard regularisation on $f$ of $\lVert f \rVert_{\rkhs}^2$.
The log-likelihood objective becomes
\[
    \frac12 \sum_{i=1}^n \left\{ \log \nu_i^\alpha + \frac{\left(y_i - \xi_i^\alpha \right)^2}{\xi_i^\alpha} \right\} + \frac{\alpha\tp K_\mathbf{z} \alpha}{2 \rho^2}
.\]
We can use backpropagation to learn the parameters $\alpha$, $\sigma$, and if we wish $\eta$, $\mathbf z$, and any kernel parameters.
The full model is illustrated in Figure \ref{fig:model-shrinkage}.
This approach allows us to directly encode uncertainty based on bag size in the objective function,
and gives probabilistic predictions.

\subsection{Bayesian Distribution Regression}
It is natural to combine the two Bayesian models above,
fully propagating uncertainty in estimation of the mean embedding and of the regression coefficients $\alpha$. Unfortunately, conjugate Bayesian inference is no longer available.
Thus, we consider a Markov Chain Monte Carlo (MCMC) sampling based approach,
and here use Hamiltonian Monte Carlo (HMC) for efficient inference,
though any MCMC-type scheme would work.
Whereas inference above used gradient descent to maximise the marginal likelihood, with the gradient calculated using automatic differentiation, here we use automatic differentiation to calculate the gradient of the joint log-likelihood and follow this gradient as we perform sampling over the parameters
we wish to infer.

We can still exploit the conjugacy of the mean shrinkage layer, obtaining
an analytic posterior over the mean embeddings.
Conditional on the mean embeddings,
we have a Bayesian linear regression model with parameters $\alpha$.
We sample this model with the NUTS HMC sampler \citep{hoffman-gelman:2012,stan-software:2014}.

\section{RELATED WORK}
As previously mentioned,
\citet{szabo2015two} provides a thorough learning-theoretic analysis of the regression model discussed in Section~\ref{sec:standard-dist-reg}.
This formalism considering a kernel method on distributions using their embedding representations, or various scalable approximations to it, has been widely applied
\citep[e.g.][]{muandet:smm,Yoshikawa2014,flaxman2015ecological,JitGreHeeEslLakSejSza2015,Lopez-paz2015,MitSejTeh2016}.
There are also several other notions of similarities on distributions in use (not necessarily falling within the framework of kernel methods and RKHSs), as well as local smoothing approaches,
mostly based on estimates of various probability metrics
\citep{Moreno2003,Jebara2004,Poczos2011,Oliva2013,poczos2013distribution,Kusner2015}.
For a partial overview, see \citet{sutherland2016scalable}.

Other related problems of learning on instances with group-level labels include
learning with label proportions \citep{quadrianto2009estimating,patrini2014almost},
ecological inference \citep{king1997solution,gelman2001models},
pointillistic pattern search \citep{Ma2015}, multiple instance learning \citep{dietterich1997solving,kuck2005learning,zhou2009multi,krummenacher2013ellipsoidal} and learning with sets \citep{zaheer2017deep}.%
\footnote{For more, also see {\scriptsize \httpurl{giorgiopatrini.org/nips15workshop}}.}

There have also been some Bayesian approaches in related contexts,
though most do not follow our setting where the label is a function of the underlying distribution rather than the observed sample set.
\citet{kuck2005learning} consider an MCMC method with group-level labels but focus on individual-level classifiers, while \cite{jackson2006improving} use hierarchical Bayesian models on both individual-level and aggregate data for ecological inference.

\citet{JitGreHeeEslLakSejSza2015} and \citet{flaxman2015ecological} quantify the uncertainty of distribution regression models by interpreting the kernel ridge regression on embeddings as Gaussian process regression. However, the former's setting has no uncertainty in the mean embeddings, while the latter's treats empirical embeddings as fixed inputs to the learning problem (as in Section \ref{section:blr}).

There has also been generic work on input uncertainty in Gaussian process regression \citep{Girard2004,Damianou2016}.
These methods could provide a framework towards allowing for second-level kernels in our models.
One could also, though, consider regression with uncertain inputs as a special case of distribution regression,
where the label is a function of the distribution's mean
and $N_i = 1$.

\section{EXPERIMENTS}
\label{section:experiments}
We will now demonstrate our various Bayesian approaches:
the mean-shrinkage pooling method with $r = k$ (\textit{shrinkage})
and with $r(x, x') = \int k(x, z) k(z, x') \nu(\mathrm d z)$ for $\nu$ proportional to a Gaussian measure (\textit{shrinkageC}),
Bayesian linear regression (\textit{BLR}),
and the full Bayesian distribution regression model with $r = k$ (\textit{BDR}).
We also compare the non-Bayesian baselines
\textit{RBF network} (Section \ref{section:baseline-model})
and \textit{freq-shrinkage},
which uses the shrinkage estimator of \citet{muandet2014kernel}
to estimate mean embeddings.
Code for our methods and to reproduce the experiments is available at
\url{https://github.com/hcllaw/bdr}.

We first demonstrate the characteristics of our models on a synthetic dataset, and then evaluate them on a real life age prediction problem. Throughout, for simplicity, we take $\mathbf{u} = \mathbf{z}$, i.e. $R = R_{\mathbf{z}} = R_{\mathbf{zz}}$, and $K_\mathbf{z} = K$~--~although $\mathbf{u}$ and $\mathbf{z}$ could be different, with $\mathbf{z}$ learnt.
Here $k$ is the standard RBF kernel.
We tune the learning rate, number of landmarks, bandwidth of the kernel and regularisation parameters on a validation set.
For BDR, we use weakly informative normal priors (possibly truncated at zero);
for other models, we learn the remaining parameters.

\subsection{Gamma Synthetic Data}\label{sec:gamma}
We create a synthetic dataset by repeatedly sampling from the following hierarchical model,
where $y_i$ is the label for the $i$th bag,
each $x^i_{j} \in \R^5$ has entries i.i.d.\ according to the given distribution,
and $\varepsilon$ is an added noise term which differs for the two experiments below:
\begin{align*}
	y_i & \sim \mbox{Uniform}(4,8) \\
    \left[x^i_{j}\right]_\ell \mid y_i  & \iid \frac{1}{y_i} \left[ \Gamma\left(\frac{y_i}{2}, \frac12 \right) \right] + \varepsilon
    \text{ for } j \in [N_i], \ell \in [5]
.\end{align*}

In these experiments,
we generate $1\,000$ bags for training, $500$ bags for a validation set for parameter tuning,
$500$ bags to use for early-stopping of the models, and $1\,000$ bags for testing.
Tuning is performed to maximize log-likelihoods for Bayesian models, MSE for non-Bayesian models.
Landmark points $\mathbf{u}$ are chosen via $k$-means (fixed across all models).
We also show results of the Bayes-optimal model, which gives true posteriors according to the data-generating process; this is the best performance any model could hope to achieve.
Our learning models, which treat the inputs as five-dimensional, fully nonparametric distributions,
are at a substantial disadvantage even in how they view the data compared to this true model.

\begin{figure}[t]
    \centering
    \includegraphics[width=.44\textwidth]{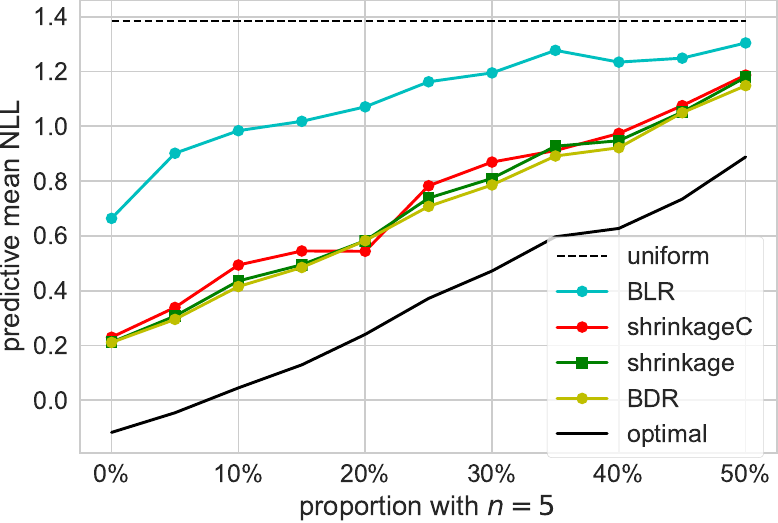}\\
    \includegraphics[width=.44\textwidth]{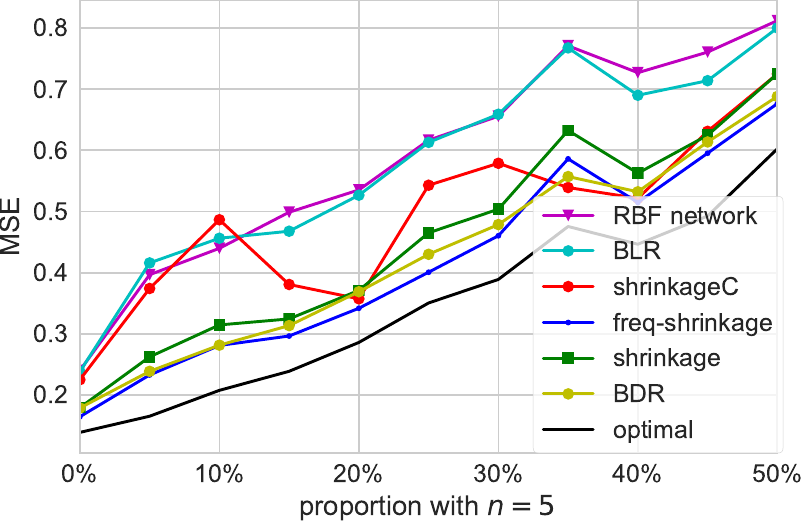}
    \caption{Top: negative log-likelihood. Bottom: mean-squared error. For context, performance of the Bayes-optimal predictor is also shown, and for NLL `uniform' shows the performance of a uniform prediction on the possible labels. For MSE, the constant overall mean label predictor achieves about 1.3.}
       \label{fig:gamma}
\end{figure}

\begin{figure*}[p]

	\centering
	\includegraphics[width=\textwidth]{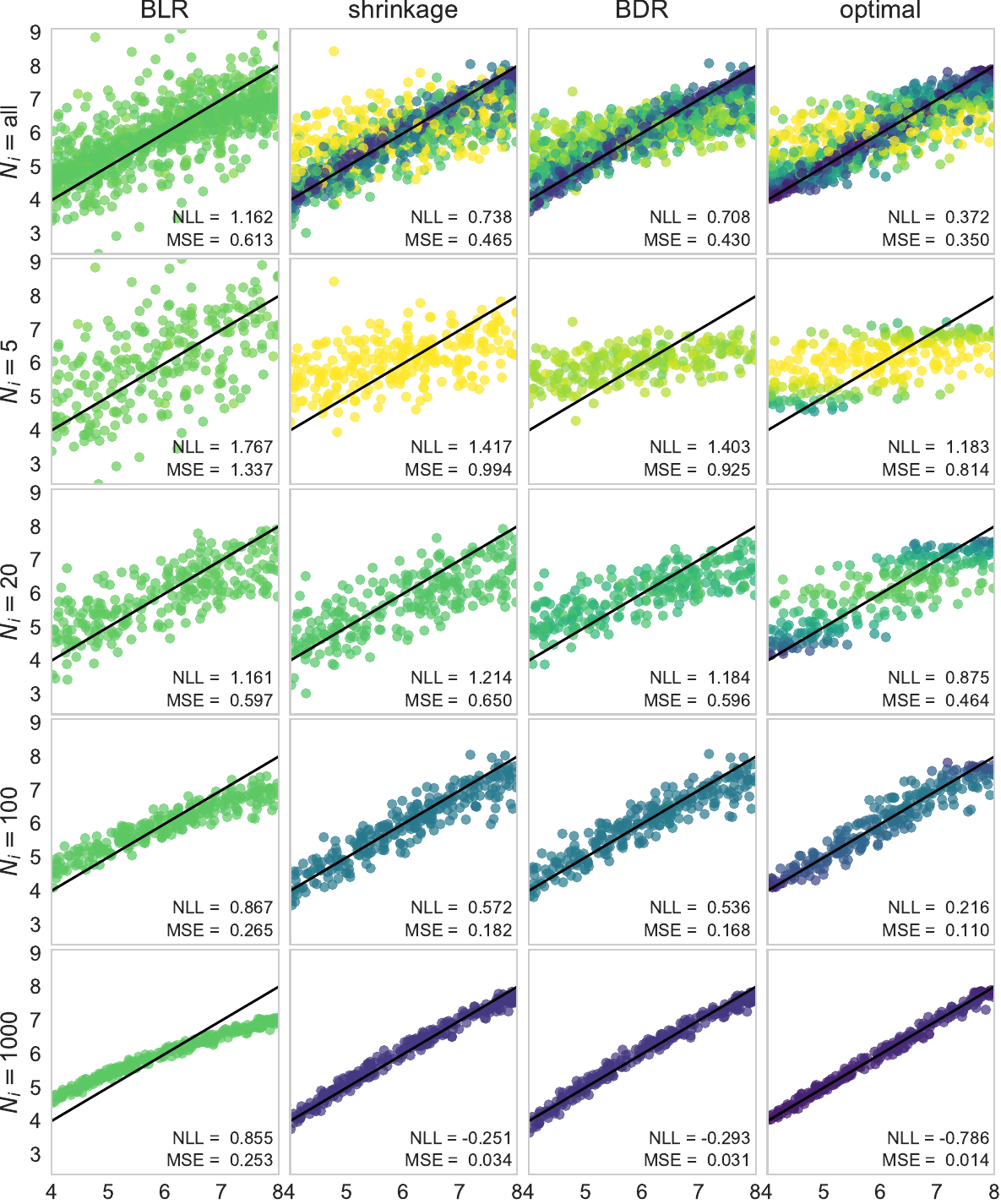}
  \\[3ex]
  \includegraphics[width=\textwidth]{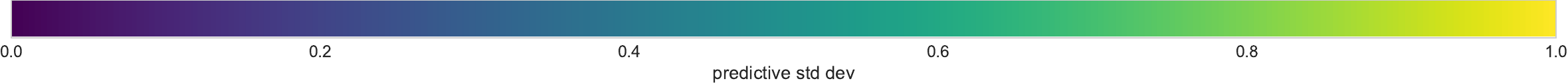}
	\caption[If there's a list of figures, we'll be sad.]{Predictions for the varying bag size experiment of Section~\ref{sec:gamma}.
		Each column corresponds to a single prediction method.
		Each point in an image represents a single bag,
		with its horizontal position the true label $y_i$,
		and its vertical position the predicted label.
		The black lines show theoretical perfect predictions.
		The rows represent different subsets of the data:
		the first row shows all bags,
		the second only bags with $N_i = 5$,
		and so on.
		Colours represent the predictive standard deviation of each point.
	}
	\label{fig:gamma-preds}

\end{figure*}

\paragraph{Varying bag size: Uncertainty in the inputs.}
In order to study the behaviour of our models with varying bag size,
we fix four sizes $N_i \in \{5, 20, 100, 1\,000\}$.
For each generated dataset,
$25\%$ of the bags have $N_i = 20$,
and $25\%$ have $N_i = 100$.
Among the other half of the data,
we vary the ratio of $N_i = 5$ and $N_i = 1\,000$ bags to demonstrate the methods' efficacy at dealing with varied bag sizes:
we let $s_5$ be the overall percentage of bags with $N_i = 5$,
ranging from $s_5 = 0$ (in which case no bags have size $N_i = 5$)
to $s_5 = 50$ (in which case $50\%$ of the overall bags have size $N_i=5$).
Here we do not add additional noise: $\varepsilon = 0$.

Results are shown in Figure \ref{fig:gamma}.
BDR and shrinkage methods, which take into account bag size uncertainty, perform well here compared to the other methods.
The full BDR model very slightly outperforms the Bayesian shrinkage models in both likelihood and in mean-squared error;
frequentist shrinkage slightly outperforms the Bayesian shrinkage models in MSE, likely because it is tuned for that metric.
We also see that the choice of $r$ affects the results; $r = k$ does somewhat better.

Figure \ref{fig:gamma-preds} demonstrates in more detail the difference between these models.
It shows test set predictions of each model on the bags of different sizes.
Here, we can see explicitly that the shrinkage and BDR models are able to take into account the bag size, with decreasing variance for larger bag sizes,
while the BLR model gives the same variance for all outputs.
Furthermore, the shrinkage and BDR models can shrink their predictions towards the mean more for smaller bags than larger ones:
this improves performance on the small bags while still allowing for good predictions on large bags, contrary to the BLR model.

\paragraph{Fixed bag size: Uncertainty in the regression model.}
The previous experiment showed the efficacy of the shrinkage estimator in our models,
but demonstrated little gain from posterior inference for regression weights $\beta$ over their MAP estimates, i.e. there is no discernible improvement of BLR over RBF network. To isolate the effect of quantifying uncertainty in the regression model, we now consider the case where there is no variation in bag size at all and normal noise is added onto the observations. In particular we take $N_i=1000$ and $\varepsilon \sim \N(0, 1)$, and sample landmarks randomly from the training set.

Results are shown in Table \ref{tab:chi}. Here, BLR or BDR outperform all other methods on all runs, highlighting that uncertainty in the regression model is also important for predictive performance.
Importantly, the BDR method performs well in this regime as well as in the previous one.

\begin{table}[ht]
\caption{Results on the fixed bag size dataset, over $10$ dataset draws (standard deviations in parentheses). BLR/BDR perform best on all runs in both metrics.} \label{tab:chi}
\begin{center}
\begin{tabular}{lll}
{\bf METHOD}  &{\bf MSE}&{\bf NLL} \\
\hline \\
    Optimal  &0.170 (0.009) & 0.401 (0.018) \\
RBF network  &0.235 (0.014) & -- \\
freq-shrinkage &0.232 (0.012) & -- \\
shrinkage &0.237 (0.014)& 0.703 (0.027)\\
shrinkageC &0.236 (0.013) & 0.700 (0.029)\\
\bf BLR & \bf 0.228 (0.012) & \bf 0.681 (0.025) \\
\bf BDR &\bf 0.227 (0.012) &\bf 0.683 (0.025)
\end{tabular}
\end{center}
\end{table}

\subsection{IMDb-WIKI: Age Estimation} \label{sec:age}
\begin{table}[t]
\caption{Results on the grouped IMDb-WIKI dataset over ten runs (standard deviations in parentheses). Here shrinkage methods perform the best across all $10$ runs.} \label{tab:face}
\begin{center}
\begin{tabular}{lll}
{\bf METHOD}  &{\bf RMSE}&{\bf NLL} \\
\hline \\
CNN             &10.25 (0.22) & 3.80 (0.034) \\
RBF network  &9.51 (0.20) & -- \\
freq-shrinkage & \bf 9.22 (0.19) & -- \\
\bf shrinkage & 9.28 (0.20) & \bf 3.54 (0.021)\\
BLR &9.55 (0.19)&3.68 (0.021)
\end{tabular}
\end{center}
\end{table}
We now demonstrate our methods on a celebrity age estimation problem, using the IMDb-WIKI database \citep{Rothe-IJCV-2016} which consists of $397\,949$ images of $19\,545$ celebrities\footnote{We used only the IMDb images, and removed some implausible images, including one of a cat and several of people with supposedly negative age, or ages of several hundred years.}, with corresponding age labels. This database was constructed by crawling IMDb for images of its most popular actors and directors, with potentially many images for each celebrity over time.
\citet{Rothe-IJCV-2016} use a convolutional neural network (CNN) with a VGG-16 architecture to perform 101-way classification, with one class corresponding to each age in $\{0, \dots, 100 \}$.

We take a different approach, and assume that we are given several images of a single individual (i.e. samples from the distribution of celebrity images), and are asked to predict their mean age based on several pictures.
For example, we have 757 images of Brad Pitt from age 27 up to 51, while we have only 13 images of Chelsea Peretti at ages 35 and 37.
Note that 22.5\% of bags have only a single image.
We obtain $19\,545$ bags, with each bag containing between $1$ and $796$ images of a particular celebrity, and the corresponding bag label calculated from the average of the age labels of the images inside each bag.

In particular, we use the representation $\varphi(x)$ learnt by the CNN in \cite{Rothe-IJCV-2016}, where $\varphi(x): \ \mathbb{R}^{256 \times 256} \to \mathbb{R}^{4096}$ maps from the pixel space of images to the CNN's last hidden layer. With these new representations, we can now treat them as inputs to our radial basis network, shrinkage (taking $r=k$ here) and BLR models. Although we could also use the full BDR model here, due to the computational time and memory required to perform proper parameter tuning, we relegate this to a later study.

We use $9\,820$ bags for training, $2\,948$ bags for early stopping, $2\,946$ for validation and $3\,928$ for testing.
Landmarks are sampled without replacement from the training set.

We repeat the experiment on $10$ different splits of the data, and report the results in Table \ref{tab:face}.
The baseline \textit{CNN} results give performance by averaging the predictive distribution from the model of \citet{Rothe-IJCV-2016} for each image of a bag; note that this model was trained on all of the images used here. From Table \ref{tab:face}, we can see that the shrinkage methods have the best performance; they outperforms all other methods in all $10$ splits of the dataset, in both metrics.
Non-Bayesian shrinkage again yields slightly better RMSEs, likely because it is tuned for that metric.
This demonstrates that modelling bag size uncertainty is vital.
\section{CONCLUSION}
Supervised learning on groups of observations using kernel mean embeddings typically disregards
sampling variability within groups.
To handle this problem, we construct Bayesian approaches to modelling kernel mean embeddings within a regression model,
and investigate advantages of uncertainty propagation within different components of the resulting distribution regression.
The ability to take into account the uncertainty in mean embedding estimates is demonstrated to be key for constructing models with good predictive performance when group sizes are highly imbalanced.
We also demonstrate that the results of a complex neural network model for age estimation can be improved by shrinkage.

Our models employ a neural network formulation to provide more expressive feature representations and learn discriminative embeddings.
Doing so makes our model easy to extend to more complicated featurisations than the simple RBF network used here.
By training with backpropagation,
or via approximate Bayesian methods such as variational inference,
we can easily `learn the kernel' within our framework,
for example fine-tuning the deep network of Section \ref{sec:age} rather than using a pre-trained model.
We can also apply our networks to structured settings,
learning regression functions on sets of images, audio, or text.
Such models naturally fit into the empirical Bayes framework.

On the other hand, we might extend our model to more Bayesian feature learning
by placing priors over the kernel hyperparameters,
building on classic work on variational approaches \citep{barber1998radial} and fully Bayesian inference \citep{andrieu2001robust} in RBF networks.
Such approaches are also possible using other featurisations, e.g.\ random Fourier features \citep[as in][]{Oliva2015}.

Future distribution regression approaches will need to account for uncertainty in observation of the distribution.
Our methods provide a strong, generic building block to do so.
\clearpage

\bibliography{biblio}

\onecolumn

\appendix
\section{Choice of \texorpdfstring{$r(\cdot,\cdot)$}{r()} to ensure \texorpdfstring{$\embeddingp \in \rkhs$}{mean embedding in the RKHS}}
\label{r_choice}
We need to choose an appropriate covariance function $r$, such that $\embeddingp \in \rkhs$, where $\embeddingp \sim \GP(0, r(\cdot,\cdot) )$. In particular, it is for infinite-dimensional RKHSs not sufficient to define $r(\cdot,\cdot) = k(\cdot, \cdot)$, as draws from this particular prior are no longer in $\rkhs$ \citep{wahba1990spline} (but see below). However, we can construct
\begin{equation}
\label{eqn:conv}
r(x, y) = \int k(x, z) k(z, y) \nu (dz)
\end{equation}
where $\nu$ is any finite measure on $\mathcal{X}$. This then ensures $\embeddingp \in \rkhs$ with probability $1$ by the nuclear dominance \citep{lukic2001stochastic, pillai2007characterizing} for any stationary kernel $k$. In particular, \cite{flaxman2016bayesian} provides details when $k$ is a squared exponential kernel defined by
$$
k(x,y) = \exp ( -\frac{1}{2} (x - y)^\top \Sigma_k^{-1} (x-y)) \quad \quad x,y \in \mathbb{R}^p
$$
and $\nu(dz) = \exp \left( - \frac{||z||^2_2}{2\ell ^2} \right) dz$, i.e. it is proportional to a Gaussian measure on $\mathbb{R}^d$, which provides $r(\cdot, \cdot)$ with a non-stationary component. In this paper, we take $\Sigma_k = \sigma^2 I_p$, where $\sigma^2$ and $\ell$ are tuning parameters, or parameters that we learn.

Here, the above holds for a general set of stationary kernels, but note that by taking a convolution of a kernel with itself, it might make the space of functions that we consider overly smooth (i.e. concentrated on a small part of $\mathcal H_k$). In this work, however, we consider only the Gaussian RBF kernel $k$. In fact, recent work \citep[Theorem 4.2]{steinwart2014convergence} actually shows that in this case, the sample paths almost surely belong to (interpolation) spaces which are infinitesimally larger than the RKHS of the Gaussian RBF kernel. This suggests that we can choose $r$ to be an RBF kernel with a length scale that is infinitesimally bigger than that of $k$; thus, in practice, taking $r=k$ would suffice and we do observe that it actually performs better (Fig. \ref{fig:gamma}).

\section{Framework for Binary Classification}
\label{app:binary}
Suppose that our labels $y_i \in \{0,1\}$, i.e. we are in a binary classification framework. Then a simple approach to accounting for uncertainty in the regression parameters is to use bayesian logistic regression, putting priors on $\beta$, i.e.
\begin{eqnarray*}
\beta & \sim & \mathcal{N}(0, \rho^2) \\
y_i &\sim & Ber(\pi_i), \text{ where } \log \left(\dfrac{\pi_i}{1-\pi_i}\right) = \beta^\top \hat{\mu}_i
\end{eqnarray*}
however for the mean shrinkage pooling model, if we use the above $y_{i}\;|\:\mu_{i}, \alpha$, we would not be able to obtain an analytical solution for $p(y_i|\mathbf{x_i}, \alpha)$. Instead we use the probit link function, as given by:
$$
Pr(y_i = 1 | \mu_{i}, \alpha)  = \Phi\left( \alpha^\top \mu_i(\mathbf{z}) \right)
$$
where $\Phi$ denotes the Cumulative Distribution Function (CDF) of a standard normal distribution, with $\mu_i(\mathbf{z}) = [\mu_i(z_1), \dots, \mu_i(z_s)]^\top$. Then as before we have
$$ \mu_{i}(\mathbf{z}) \mid \mathbf{x_i} \sim \mathcal{N}\left( M_i, C_i\right)$$
with $M_i$ and $C_i$ as defined in section \ref{section:mean-shrinkage}. Hence, as before
\begin{eqnarray*}
Pr(y_{i} = 1|{\bf x}_{i},\alpha) & = & \int Pr(y_{i} = 1|\mu_{i},\alpha)p(\mu_{i}(\mathbf{z} )|{\bf x}_{i})d\mu_{i}(\mathbf{z} )\\
 & = & c\int \Phi(\alpha^{\top}\mu_i(\mathbf{z} )) \exp \{ -\frac{1}{2} (\mu_i(\mathbf{z} ) - M_i) ^\top C_i ^ {-1} ( \mu_i(\mathbf{z} ) - M_i ) \} d\mu_i(\mathbf{z} ) \\
( \text{with }l_i = \mu_i(\mathbf{z} ) - M_i) & = & c\int \Phi(\alpha^{\top}(l_i + M_i)) \exp \{ -\frac{1}{2} (l_i) ^\top C_i ^ {-1} ( l_i) \} dl_i \\
& = & Pr(Y\leq \alpha^\top (l_i + M_i) )
\end{eqnarray*}
Note here $Y \sim \mathcal{N}(0,1)$ and $l_i\sim \mathcal{N}(0, \Sigma_i)$ Then expanding and rearranging
\begin{eqnarray*}
Pr(y_{i} = 1|{\bf x}_{i},\alpha) & = & Pr(Y - \alpha^\top l_i \leq \alpha^\top M_i )
\end{eqnarray*}
Note that since $Y$ and $l_i$ independent normal r.v., $Y - \alpha^\top l_i \sim \mathcal{N}(0, 1+\alpha^\top C_i \alpha^\top)$. Let $T$ be standard normal, then we have:
\begin{eqnarray*}
Pr(y_{i} = 1|{\bf x}_{i},\alpha) & = & Pr( \sqrt{1+\alpha^\top C_i \alpha} \ T \leq \alpha^\top M_i ) \\
& = & Pr(  T \leq \frac{\alpha^\top M_i}{ \sqrt{1+\alpha^\top C_i \alpha}} ) \\
& = & \Phi \left( \frac{\alpha^\top M_i}{ \sqrt{1+\alpha^\top C_i \alpha}} \right)
\end{eqnarray*}
Hence, we also have:
\begin{eqnarray*}
Pr(y_{i} = 0|{\bf x}_{i},\alpha) & = & 1- \Phi \left( \frac{\alpha^\top M_i}{ \sqrt{1+\alpha^\top C_i \alpha}} \right)
\end{eqnarray*}
Now placing the prior $\alpha \sim \mathcal{N}(0, \rho^2 K_{\mathbf{z}}^{-1})$, we have the following MAP objective:
\begin{eqnarray*}
J(\alpha) & = & \log\left[p(\alpha)\prod_{i=1}^{n}p(y_{i}|{\bf x}_{i},\alpha) \right]\\
 & = & \sum_{i=1}^{n} (1-y_i) \log ( 1-  \Phi \left( \frac{\alpha^\top M_i}{ \sqrt{1+\alpha^\top C_i \alpha}} \right) ) \\
 && + y_i \log (\Phi \left( \frac{\alpha^\top M_i}{ \sqrt{1+\alpha^\top C_i \alpha}} \right)) +\frac{1}{\rho^{2}}\alpha^\top K_\mathbf{z} \alpha
 \end{eqnarray*}
Since we have an analytical solution for $Pr(y_{i} = 0|{\bf x}_{i},\alpha)$, we can also use this in HMC for BDR.
\section{Some more intuition on the shrinkage estimator}
\label{app:shrink}
In this section, we provide some intuition behind the shrinkage estimator in section \ref{section:mean-shrinkage}. Here, for simplicity, we choose $\Sigma_i = \tau^2 I$ for all bag $i$, and $m_0 = 0$, and consider the case where $\mathbf{z} = \mathbf{u}$, i.e. $R = R_\mathbf{z} = R_{\mathbf{zz}}$. We can then see that if $R$ has eigendecomposition $U \Lambda U^T$,
with $\Lambda = \diag(\lambda_k)$, the posterior mean is
\[
	U \diag\left( \frac{\lambda_k}{\lambda_k + \tau^2 / N_i} \right) U^T (\hat{\mu}_i)
,\]
so that large eigenvalues, $\lambda_k \gg \tau^2 / N_i$, are essentially unchanged,
while small eigenvalues, $\lambda_k \ll \tau^2 / N_i$, are shrunk towards 0.
Likewise, the posterior variance is
\[
	U \diag\!\left( \lambda_k - \frac{\lambda_k^2}{\lambda_k + \frac{\tau^2}{N_i}} \right) U^T
	= U \diag\!\left( \frac{1}{\frac{N_i}{\tau^2} + \frac{1}{\lambda_k}} \right) U^T
;\]
its eigenvalues also decrease as $N_i / \tau^2$ increases.
\section{Alternative Motivation for choice of \texorpdfstring{$f$}{f}}
\label{app:representer}
Here we provide an alternative motivation for the choice of $f = \sum^k_{s=1} \alpha_s k(\cdot, z_s)$. First, consider the following Bayesian model with a linear kernel $K$ on $\mu_i$, where $f: \mathcal{H}_k \rightarrow \mathbb{R}$:
\begin{eqnarray*}
y_{i}\;|\:\mu_{i}, f & \sim & \mathcal{N}\left(f(\mu_i),\sigma^{2}\right).
\end{eqnarray*}
Now considering the log-likelihood of $\{\mu,Y\} = \{\mu_i, y_i\}^n_{i=1}$ (supposing we have these exact embeddings), we obtain:
$$
\log p (Y|\mu,f) = \sum^n_{i=1}-\frac{1}{2 \sigma^2} (y_i - f(\mu_i) )^2
$$
To avoid over-fitting, we place a Gaussian prior on $f$, i.e. $- \log p(f) = \lambda  || f ||_{\mathcal{H}_k} + c$.
Minimizing the negative log-likelihood  over $f \in \mathcal{H}_k$, we have:
$$
f^* = \text{argmin}_{f \in \mathcal{H}_k} \sum^n_{i=1}\frac{1}{2 \sigma^2} (y_i - f(\mu_i) )^2 +  \lambda  || f ||_{\mathcal{H}_k}
$$
Now this is in the form of an empirical risk minimisation problem. Hence using the representer theorem \citep{representer}, we have that:
\begin{eqnarray*}
f = \sum_{j = 1}^n \gamma_j K ( . , \mu_j )
  \end{eqnarray*}
i.e. we have a finite-dimensional problem to solve. Thus since $K$ is a linear kernel:
\begin{eqnarray*}
y_{i}\;|\:\mu_{i}, \{\mu_j\}_{j=1}^n,\gamma & \sim & \mathcal{N}\left( \sum_{j=1}^n \gamma_j \langle \mu_i, \mu_j\rangle_{\mathcal{H}_k},\sigma^{2}\right).
\end{eqnarray*}
where $\langle \mu_i, \mu_j\rangle_{\mathcal{H}_k}$ can be thought of as the similarity between distributions.

Now we have the same $\GP$ posterior as in Section \ref{section:mean-shrinkage}, and we would like to compute $p(y_i | \mathbf{x_i}, \gamma)$.
This suggests we need to integrate out $\mu_1$, \dots $\mu_n$.
But it is unclear how to perform this integration, since the $\mu_i$ follow Gaussian process distributions.
\iffalse
so we need to compute the following ($\mu_{i}$ index appears once) CHECK THIS notation..... :
\begin{eqnarray*}
p(y_{i}|{\bf x}_{i},\gamma) & = & \int p(y_{i}|\mu_{i},  \{\mu_j\}_{j=1}^n,\gamma)p(\mu_{i}|{\bf x}_{i}) p(\mu_{1}|\mathbf{x_1}) \dots p(\mu_{n}|\mathbf{x_n}) d\mu_{i} d\mu_1 \dots d\mu_{n}
\end{eqnarray*}
which is a integral over a Gaussian Process, which is complex.
\fi
 Hence we can take an approximation to $f$, i.e. $f = \sum^k_{s=1} \alpha_s k(\cdot, z_s)$, which would essentially give us a dual method with a sparse approximation to $f$.

\end{document}